\documentclass[letterpaper, 10 pt, conference]{ieeeconf}  

\IEEEoverridecommandlockouts                              

\usepackage{graphicx}
\usepackage{amsmath} 
\usepackage{amssymb}  
\usepackage{xcolor}
\usepackage{booktabs}
\usepackage{censor}

\usepackage{microtype}

\title{\LARGE \bf
STEMbot: A Compliant Robot for Under-Canopy Plant Navigation
}

\author{
Zachary Charlick$^{1}$,
Nilay Roy Choudhury$^{1}$,
Haoyu Ma$^{1}$,
Xiaonan Huang$^{1}$,
and Dmitry Berenson$^{1}$%
\thanks{$^{1}$Robotics Department, University of Michigan, Ann Arbor, MI, USA.}%
\thanks{$^{2}$This work was supported in part by the Office of Naval Research Grant N00014-24-1-2036, NSF grants IIS-2113401 and IIS-2220876, and the National Robotics Initiative, project award no. 2026-67021-46039, from the U.S. Department of Agriculture’s National Institute of Food and Agriculture.}}%

\begin{document}

\maketitle
\thispagestyle{empty}
\pagestyle{empty}


\begin{abstract}
The scalability of organic agriculture is partially limited by the labor costs associated with monitoring for pests. While drones and rovers are well-suited for agricultural monitoring from above or next to plants, many pests live on the underside of leaves or on plant stems, making them detectable only \textit{after} they have caused significant damage. To enable early pest detection we present STEMbot, a miniature climbing robot system designed for autonomous navigation \textit{under plant canopies}. Unlike existing climbing platforms that lack on-board perception or are restricted to unbranched vertical trunks, STEMbot integrates a fully geometric PIN-SLAM pipeline with a semantic OcTree to achieve robust localization and mapping while climbing the plant. To plan STEMbot's motion we propose a manifold-constrained A* planner along with ray-tracing goal specification to enable branch-aware traversal and the inspection of occluded targets. We validate our system through hardware experiments, demonstrating reliable traversal of stems ranging from 7--33\,mm and autonomous navigation across four distinct plant specimens. Quantitative evaluations show that our system achieves high-fidelity geometric reconstructions with an average Chamfer distance of less than 1\,cm relative to an offline photogrammetry baseline, confirming that STEMbot maintains the globally consistent odometry needed for autonomous navigation.
\end{abstract}


\section{INTRODUCTION}

Existing methods for insect pest detection in organic vegetable farming---e.g. Integrated Pest Management (IPM) \cite{USDA_IPM}---involve frequent manual inspections and pest-specific trapping at key times during a pest's lifecycle. These methods require significant labor and expertise, and the associated expense limits the scalability of organic farming. While drones and rovers are able to monitor plants from a distance, many pests live on the underside of leaves or on plant stems, making them detectable from a distance only \textit{after} they have caused significant damage. In this systems paper, we propose STEMbot, a novel climbing robot design and software framework for autonomous navigation \textit{under plant canopies}. By navigating on stems and branches, this robot is able to work around occlusions so that it can inspect specific areas of the plant where a target pest is likely to reside\textsuperscript{*}.

\begingroup
\renewcommand{\thefootnote}{\fnsymbol{footnote}}
\footnotetext[1]{Specifically, this robot is intended for high-value crops with climbable stems such as tomatoes, peppers, and cucurbits to detect pests such as tomato hornworms, cucumber beetles, and aphids.}
\endgroup

Automating plant navigation requires robust perception, that is able to handle unstructured geometry and extreme occlusion. 
Furthermore, the repetitive textures and largely monochromatic appearance of plant foliage introduce substantial perceptual aliasing, which degrades the performance of many perception methods.

\begin{figure}[t]
    \centering
    \includegraphics[width=\linewidth]{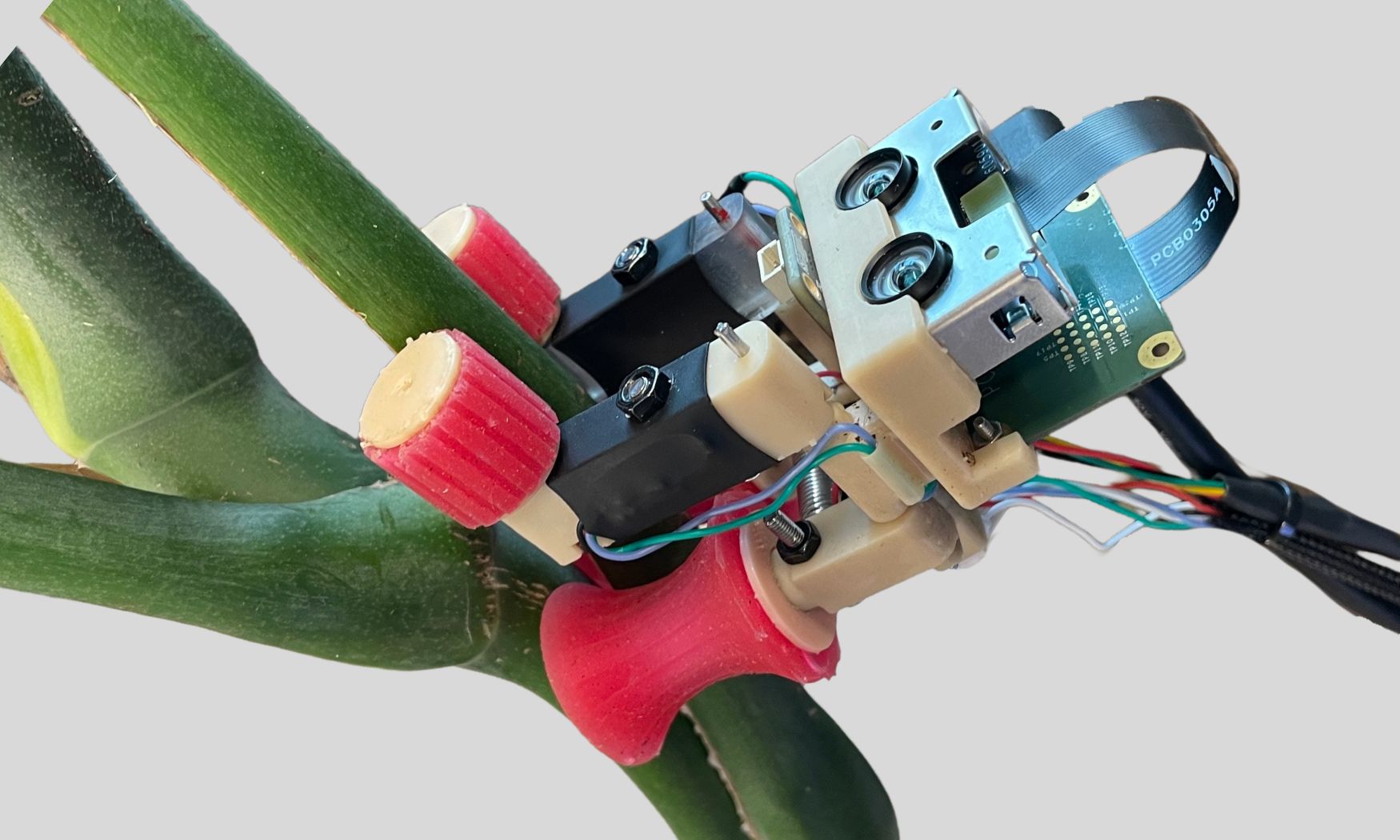}
    \vspace{-20pt}
    \label{fig:hero_fig}
    \caption{STEMbot navigating under a plant canopy. The robot integrates perception, motion planning, and specialized hardware to maintain stable contact with the stem manifold, enabling autonomous localization and semantic mapping.}
    \vspace{-18pt}
\end{figure}

While many climbing robots have been developed, most are designed for industrial inspection tasks, such as robots that climb and inspect pipes or infrastructure surfaces \cite{halder2023robots, chattopadhyay2018locomotion, hernando2019romerin, choudhury2026slipstream}. Tree-climbing robots have also been explored, but many of these systems are designed to fully-enclose a single trunk, such as a coconut tree, where the absence of branches significantly simplifies navigation. A few more flexible tree-climbing platforms exist, such as Treebot \cite{treebot}; however, these systems typically are much larger than is practical for our application and lack onboard vision or odometry. As a result, they can only perform local, reactive steps and are unable to plan long paths or map their environment.

We present the STEMbot climbing robot, a novel hardware design with perception and planning methods appropriate for under-canopy navigation. The contributions of this paper are:

\begin{itemize}
    \item A plant-climbing robot capable of traversing stems ranging from 7\,mm to 33\,mm in diameter, transitioning onto branches, and maintaining contact while inverted.
    
    \item A semantic OcTree mapping pipeline that combines PIN-SLAM, SAM, and CLIP to provide millimeter-resolution semantic reconstruction and odometry in plant environments.
    
    \item A manifold-constrained A* motion planner for branch-aware navigation along plant structures.

    \item Experiments on physical plants in a lab setting, demonstrating navigation and reconstruction.
\end{itemize}

\section{RELATED WORK}

\subsection{Climbing Robots and Tree Traversal}

Climbing robots can be broadly categorized by their attachment mechanisms. Industrial platforms often utilize magnetic, vacuum, or microspine adhesion (e.g., RiSE \cite{spenko2008rise}) to traverse man-made surfaces. However, these mechanisms struggle with the irregular, porous, and non-ferromagnetic geometries of living plants. Traction-based designs like Amaran \cite{megalingam2021amaran} and Climbot \cite{guan2016climbot} employ encirclement strategies. While effective for uniform poles, lateral branches and protrusions, preventing navigation through complex canopies.

To our knowledge, Treebot \cite{treebot} is the only other platform to demonstrate a trunk-to-branch transition on a real tree. This 600\,g continuum robot utilized two omnidirectional claw grippers with needle-tipped phalanges to navigate from a 280\,mm trunk to a 118\,mm branch via tactile-based reactive planning. Treebot’s needle-based grip risks damaging bark and is limited to branches larger than $118\,\text{mm}$. Branch Bot \cite{rozenlevy2021branchbot} utilized passive elastic grip for rod-like structures but remained tethered and limited to single-segment traversal. Our system extends these capabilities by utilizing high-friction compliant wheels for the safe traversal of delicate stems as small as $7\,\text{mm}$. Our approach integrates a full perception and SLAM stack with a global motion planner to autonomously navigate or avoid branches.

\subsection{SLAM in Challenging or Low-Texture Environments}
Traditional vision-based SLAM frameworks, including feature-based methods like \cite{campos2021orb} and direct methods like Direct Sparse Odometry (DSO) \cite{engel2017direct}, rely on visual correspondences that often fail in plant canopies due to repetitive textures and monochromatic color palettes, which induce significant perceptual aliasing. While learning-based methods like DROID-SLAM \cite{teed2021droid} or VGGT-SLAM \cite{wang2025vggt} offer increased robustness in natural scenes, they often degrade when encountering the extreme viewpoints or scale changes typical of close-quarters stem traversal. To address these challenges, we adopt PIN-SLAM \cite{pan2024pin}, a geometric framework that utilizes depth-based features rather than visual appearance. PIN-SLAM optimizes local signed distance functions (SDFs) anchored by a sparse set of elastic neural points; these act as geometric anchors for bundle adjustment and global loop closure, providing stability without requiring photometric consistency.

\subsection{Semantic Mapping and Octree Representations}
While occupancy grids traditionally discretize environments into probabilistic cells \cite{moravec1985high}, 3D voxel implementations often incur prohibitive memory overhead; hierarchical spatial subdivision, such as OctoMap \cite{hornung2013octomap}, mitigates this by allocating resolution dynamically. Recent advancements have further integrated semantic data into these structures using Bayesian log-odds updates \cite{thrun2002probabilistic, asgharivaskasi2023semantic} to maintain probabilistic class distributions. Furthermore, foundation models like Segment Anything (SAM) \cite{kirillov2023segment} and CLIP \cite{radford2021learningtransferablevisualmodels} have enabled open-world perception through semantic masking and vision-language alignment. Recent architectures like DINOv3 \cite{simeoni2025dinov3} and Florence-2 \cite{xiao2024florence} further consolidate these capabilities into unified, zero-shot frameworks for dense semantic classification and grounding. These models provide the building blocks for constructing semantic global representations in unstructured environments.

\subsection{Motion Planning on Manifolds}

Motion planning under task constraints often involves navigating lower-dimensional manifolds embedded within a higher-dimensional state space \cite{berenson2009manipulation}. Current state-of-the-art approaches typically utilize projection-based sampling to pull configurations back onto the constraint surface \cite{kingston2018sampling} or leverage manifold samples to approximate the valid configuration space \cite{salzman2013motion}. In this work, we simplify the manifold-constrained problem by abstracting the plant morphology into a discrete semantic voxel grid. Rather than utilizing expensive numerical optimization, we employ a nearest-neighbor projection onto the traversable voxel centroids. To ensure kinematic feasibility, we utilize an A* search-based framework over a state lattice of discrete action primitives \cite{pivtoraiko2009differentially}. This approach leverages a library of feasible motions to explore the configuration space while naturally respecting physical constraints \cite{howard2008state}.

\section{SYSTEM}

\begin{figure*}[t]
    \centering
    \includegraphics[width=\linewidth]{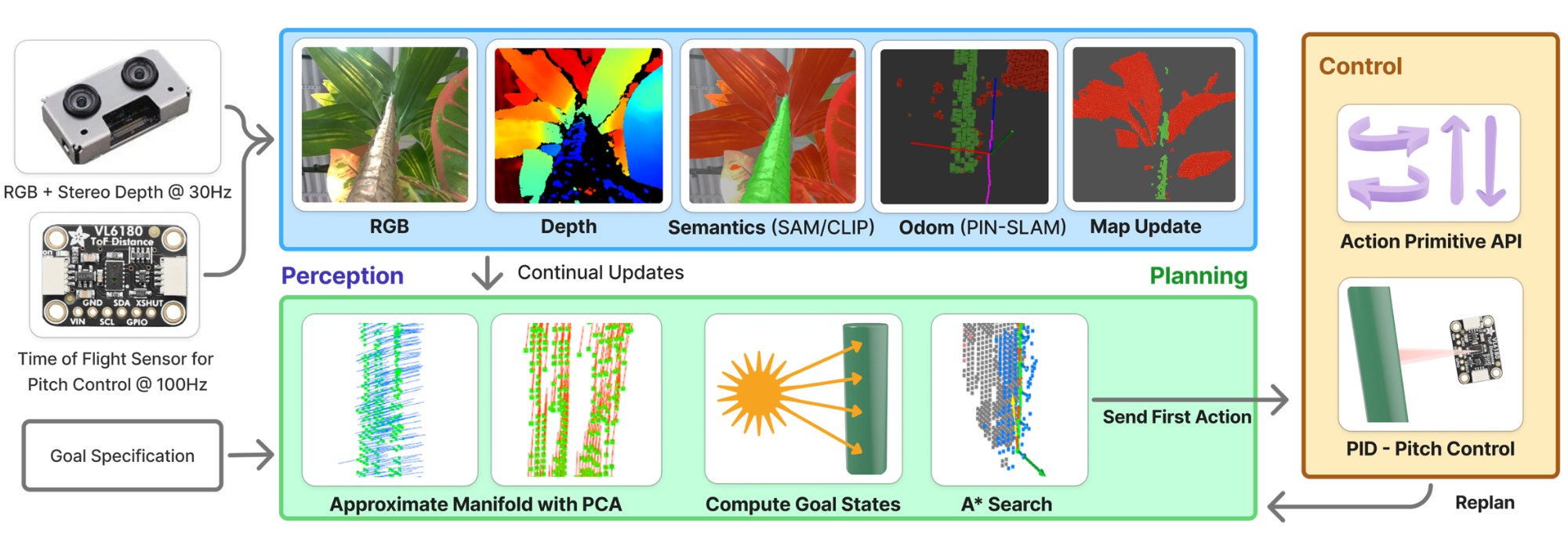}
    \vspace{-20pt}
    \caption{System architecture and data flow for the plant-climbing robot. The pipeline is divided into three primary subsystems: (Left) Sensing and Perception, which processes multi-rate stereo and ToF data through PIN-SLAM and SAM/CLIP for semantic OcTree generation; (Center) Motion Planning, where manifold estimation and A* search compute trajectories using a receding horizon framework; and (Right) Control, which executes discrete action primitives via a closed-loop PID pitch controller to generate motor commands.}
    \label{fig:system_pipeline}
    \vspace{-15pt}
\end{figure*}

Our system, illustrated in Fig.~\ref{fig:system_pipeline}, integrates novel hardware design with a perception-driven planning stack designed specifically for manifold-constrained environments. The system comprises three primary subsystems: hardware, which utilizes a spring-loaded four-bar linkage and high-friction compliant wheels to maintain stable contact with varying stem diameters; perception, which fuses geometric PIN-SLAM with foundation-model-based semantic segmentation (SAM and CLIP) to generate a probabilistic semantic OcTree; and motion planning, which employs a manifold-constrained A* search and a receding horizon framework to facilitate branch-aware traversal. STEMbot is tethered for both power and computation. An NVIDIA RTX 4080 GPU  supports perception and motion planning, while an Arduino Nano performs controls PID and low-level motor commands.

\subsection{Hardware: Actuation and Sensing}

\subsubsection{Actuation}

\begin{figure}[thpb]
    \vspace{-10pt}
    \centering
    \includegraphics[width=\linewidth]{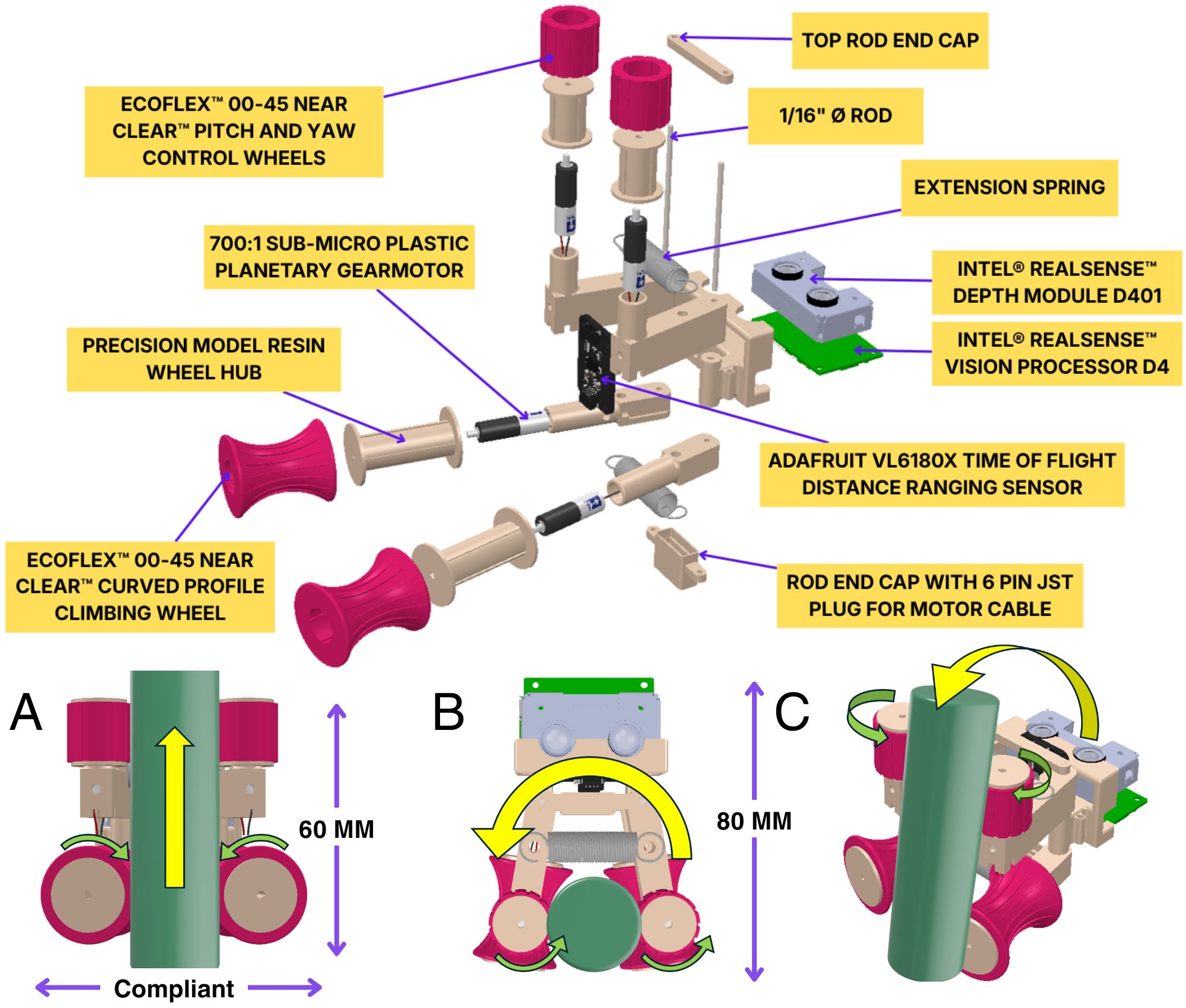}
    \vspace{-20pt}
    \caption{System hardware and locomotion modes. The exploded view (top) details the perception payload and dual-motor drive assembly; note the \textbf{concave wheel-groove geometry} designed to maximize contact with the stem manifold. The coordinated motor control (bottom) enables three motion primitives: (A)~vertical traversal, (B)~yaw rotation for circumferential positioning, and (C)~pitch adjustment.}
    \label{fig:hardware_combined}
    \vspace{-10pt}
\end{figure}

The mechanical platform, illustrated in Fig.~\ref{fig:hardware_combined}, has an untethered mass of 67 grams. STEMbot uses two pairs of Pololu 700:1 sub-micro planetary gearmotors (Item No. 2359), each providing a high-ratio reduction optimized for torque (up to 900 g-cm) climbing on stems with variable diameters. The planetary gearbox minimizes backlash and provides a compact form factor, allowing the robot to climb, rotate, and adjust its pitch.

Each motor pair drives a wheel hub printed in Precision Model Resin. Attached are compliant wheels manufactured by casting silicone molds with near-clear EcoFlex 00-45 elastomer material (Smooth-On, Inc, Shore hardness 00–45). The distinctive curved wheel profile and central groove (Fig.~\ref{fig:hardware_combined}, lower panel) are designed to grip and stabilize the robot as it traverses the stem. The high-friction, compliant material conforms to stem surface irregularities, distributing contact pressure and preventing slippage during vertical ascent and pitch maneuvers. 

The coordinated control of the two motor pairs enables three independent motion primitives—vertical traversal, yaw rotation, and pitch adjustment—as illustrated in Fig.~\ref{fig:hardware_combined}. The four-bar linkage and passive spring clamping maintain stable stem contact throughout all three modes.

\subsubsection{Sensing and Feedback Control}
\label{sec:control}

The sensor configuration includes an Intel RealSense Depth Module D401 and an Adafruit VL6180X Time-of-Flight (ToF) sensor. The stereo camera facilitates geometric SLAM and mapping, while the ToF sensor indirectly observes the robot's pitch relative to the plant stem. To prevent the robot from tipping backward and losing contact between the upper drive wheels and the stem, a closed-loop proportional-integral-derivative (PID) controller tracks a target setpoint $d_{ToF}$ by issuing corrective pitch and yaw velocity commands based on ToF distance feedback at 90Hz. The controller maps distance error to motor speed commands via tuned gain parameters (Kp = 20, Kd = 5, Ki = 0.1) optimized to balance responsiveness and stability.

\subsection{Perception}

\begin{figure}[thpb]
    \centering
    \includegraphics[width=\linewidth]{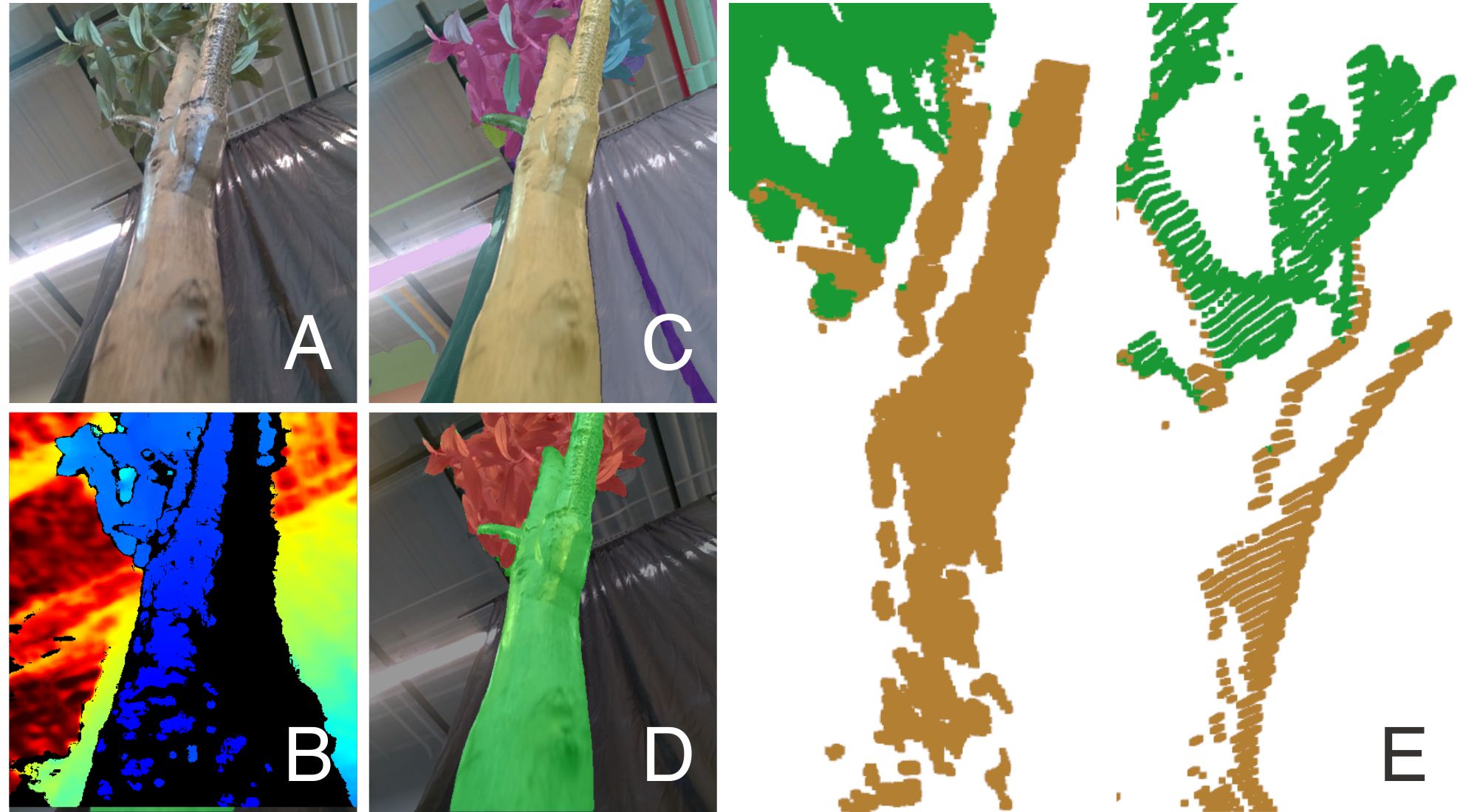}
    \caption{Overview of the semantic manifold perception pipeline. (A-B) RGB and colorized depth input from the RealSense D405. (C-D) Initial SAM segmentation refined via CLIP ($p > 0.90$) to identify traversable stems (green) and non-traversable leaves (red). (E) Semantic point cloud generated by projecting mask-specific depth values into 3D and registering them via PIN-SLAM odometry.}
    \label{fig:perception_pipeline}
\end{figure}

Figure~\ref{fig:perception_pipeline} provides an overview of the perception pipeline for semantic manifold estimation. The pipeline leverages an Intel RealSense D405 stereo camera (interfaced with a dedicated vision processor) to provide per-pixel depth computation at 30 Hz. To maintain real-time performance on our compute-constrained platform, odometry is processed at 3.75 Hz by subsampling every eighth frame. For each selected frame, the depth image (Fig.~\ref{fig:perception_pipeline}B) is projected into a 3D point cloud and registered using PIN-SLAM \cite{pan2024pin}.

\subsubsection{Geometric Registration}

PIN-SLAM is a fully geometric method \cite{cadena2017past} that utilizes depth-based features for registration. Agricultural environments present a specific challenge for photometric methods; they are often visually homogeneous, dominated by a largely homogeneous color palette and repetitive patterns. This lack of visual distinctiveness leads to significant perceptual aliasing, where different parts of the plant appear identical. While symmetric environments like tunnels or hallways frequently induce geometric degeneracy, the intricate and non-uniform morphology of plant stems provides dense spatial features. PIN-SLAM requires an update rate of 10–15 Hz to remain within the registration convergence basin. Operating at 5 Hz on an RTX 4080 GPU, we utilize a discontinuous strategy: 300 ms of motion is followed by a 1 s stationary interval for pose synchronization and concurrent motion planning.

\subsubsection{Semantic Segmentation and Classification}

The global map is updated at a frequency of 1 Hz. During each update, the system captures the latest odometry pose and RGB image (Fig.~\ref{fig:perception_pipeline}A) to perform semantic segmentation via the Segment Anything Model (SAM) \cite{kirillov2023segment}. The resulting semantic masks (Fig.~\ref{fig:perception_pipeline}C) are classified using Contrastive Language-Image Pre-Training (CLIP) \cite{radford2021learningtransferablevisualmodels} against a closed vocabulary consisting of $\{leaves, trunk, sky, light, wall, curtain, grate\}$. While the latter five classes serve primarily to filter irrelevant points, high-confidence masks ($p > 0.90$) for ``trunk'' and ``leaves'' are used to create corresponding semantic masks (Fig.~\ref{fig:perception_pipeline}D).

\subsubsection{Probabilistic Mapping}

Depth observations corresponding to the high-confidence semantic masks are back-projected into 3D space to generate semantic point clouds (Fig.~\ref{fig:perception_pipeline}E). These local coordinates are then transformed into a consistent global frame using the odometry provided by PIN-SLAM. Semantic clouds are integrated into an OcTree with Bayesian probabilistic updates to log-odds vectors \cite{asgharivaskasi2023semantic}. A softmax operation yields the probability of each voxel belonging to the traversable, non-traversable, or free-space classes.

\subsubsection{Local Geometry Estimation}
We extract the centroid of each "stem" voxel from the semantic OcTree to form a downsampled point cloud, indexed in a $k$-d tree for efficient spatial queries. We estimate the surface normal and branch direction at every point using Principal Component Analysis (PCA) over varying neighborhood scales. For each point, the surface normal $\mathbf{n}$ is computed as the smallest principal component of neighbors within a local radius $r_{normal}$, where the local patch is approximately planar. The branch heading $\mathbf{b}$ is identified as the primary principal component (the axis of maximum variance) by utilizing a larger radius $r_{branch}$ sufficient to capture the stem's longitudinal axis. Geometric estimation is performed at each motion planning iteration.

\subsection{Motion Planning}

The motion planning subsystem computes trajectories that enable the robot to navigate through a plant cannopy to reach either a specific state or vantage point for observation. We first define the formal problem and notation before describing our search-based solution.

\subsubsection{Problem Statement}

\begin{figure}[thpb]
    \centering
    \includegraphics[width=\linewidth]{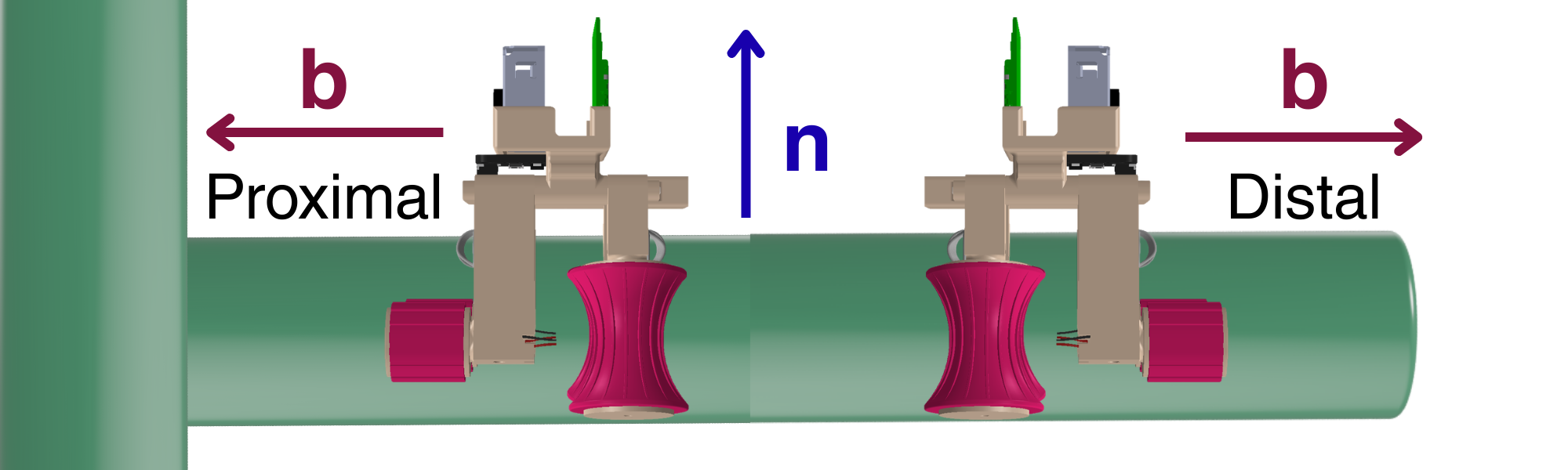}
    \caption{Geometric constraints on the branch heading vector $\mathbf{b}$. The robot's wheel geometry restricts the heading relative to the primary stem axis.}
    \label{fig:b_constraints}
\end{figure}

Let the plant environment be represented by a semantic point cloud $\mathcal{C}$, segmented and voxelized into traversable, non-traversable, and free-space voxels. We define the robot state as the tuple $s = \{\mathbf{p}, \mathbf{n}, \mathbf{b}\}$, where $\mathbf{p} \in \mathbb{R}^3$ is the 3D position of the nearest traversable voxel centroid, $\mathbf{n} \in \mathbb{R}^3$ is the surface normal at $\mathbf{p}$, and $\mathbf{b} \in \mathbb{R}^3$ is the branch heading vector at $\mathbf{p}$, constrained to be either proximal or distal relative to the stem axis (Fig.~\ref{fig:b_constraints}). For notational convenience, we will sometimes refer to  the components of a state by its subscript; e.g. to refer to the surface normal at $s_t$, we will use the notation $\mathbf{n}_t$. Together, $\mathbf{n}$ and $\mathbf{b}$ define a locally linear approximation of the \textit{plant manifold}—a 2D surface embedded in $\mathbb{R}^3$.

\begin{figure}[thpb]
    \centering
    \includegraphics[width=\linewidth]{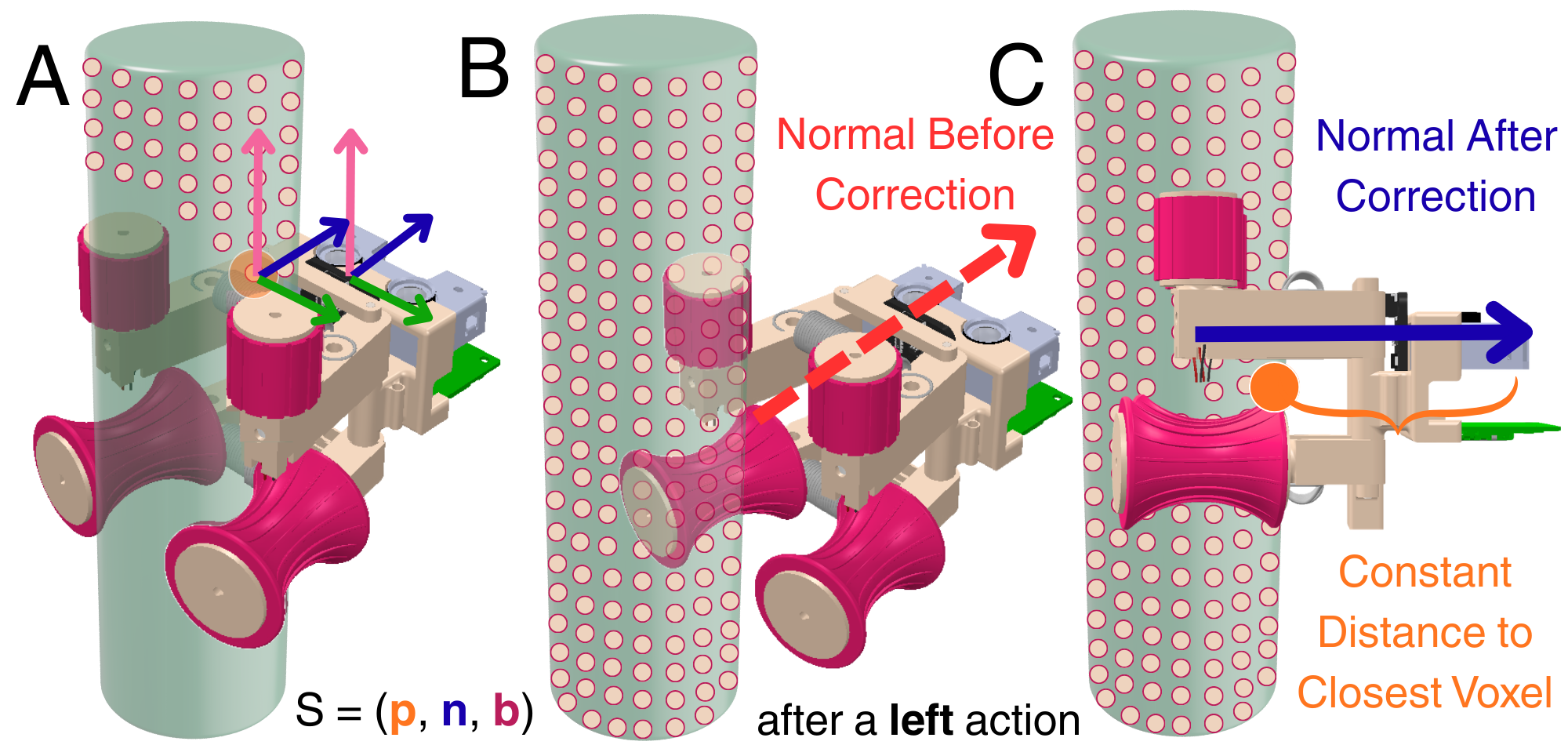}
    \caption{Discrete state representation and manifold projection for motion planning. (A) Initial robot state defined on the voxelized manifold, where the 3D position $\mathbf{p}$ is mapped to the nearest traversable voxel centroid. (B) The robot executes a discrete motion primitive (e.g., a circumferential rotation), resulting in a candidate state that deviates from the manifold surface. (C) The candidate position is re-projected onto the surface using a distance threshold to the nearest neighbor. The local geometric features—branch heading $\mathbf{b}$ and surface normal $\mathbf{n}$—are subsequently re-estimated and corrected based on the new projected manifold position.}
    \label{fig:motion_model}
    \vspace{-20pt}
\end{figure}

Given an initial state $s_0$ and a goal condition $G$, we compute a sequence of action primitives $\mathcal{A} = \{a_1, a_2, \dots, a_k\} $
that transitions the robot to a state $s \in G$ while maintaining continuous contact with the traversable manifold.

\subsubsection{Assumptions}

To ensure tractability for onboard computation, we make the following assumptions:

\begin{enumerate}
    \item \textbf{Static Environment:} The plant geometry remains fixed during the planning and execution cycle.
    \item \textbf{Local Manifold Approximation:} Within the robot’s immediate footprint, the stem is modeled as a cylindrical manifold. This reduces the local configuration space to two degrees of freedom: longitudinal translation and circumferential rotation.
    \item \textbf{Low-Level Control:} A low-level PID controller regulates robot pitch using ToF feedback (Section~\ref{sec:control}), ensuring that the branch vector $\mathbf{b}$ closely aligns with the robot heading.
\end{enumerate}

\subsubsection{Planning Challenges}

This planning problem presents three primary challenges:

\begin{enumerate}
    \item \textbf{Kinematic Constraints:} The robot must align with the stem axis, rendering many spatially proximal states unreachable.
    \item \textbf{Branch Switching Sensitivity:} Executing a branch transition requires precise geometric alignment, making the connectivity of the search graph highly sensitive to local manifold estimation errors.
\end{enumerate}

\subsubsection{Graph Search Formulation}

We use A* search over the discrete state space. Given an initial state $s_0$, adjacent states are generated by a successor transition function constrained by the robot's kinematics.

We define the transition function $s_{t+1} = f(s_t, u)$, where the action $u$ results in a naive displacement of the position $\mathbf{p}_t$. Given the discrete action $u \in \{\text{UP, DOWN, LEFT, RIGHT}\}$, the updated position is:

\vspace{-15pt}
\begin{equation}
\mathbf{p}_{t+1} = 
\begin{cases} 
\mathbf{p}_t + \delta_{linear}\mathbf{b}_t & \text{if } u = \text{UP} \\
\mathbf{p}_t - \delta_{linear}\mathbf{b}_t & \text{if } u = \text{DOWN} \\
\mathbf{p}_t + \delta_{angular}(\mathbf{b}_t \times \mathbf{n}_t) & \text{if } u = \text{LEFT} \\
\mathbf{p}_t - \delta_{angular}(\mathbf{b}_t \times \mathbf{n}_t) & \text{if } u = \text{RIGHT}
\end{cases}
\label{eq:naive_update}
\end{equation}
 \vspace{-10pt}
 
\subsubsection{Manifold Projection}

Applying these equations directly may cause the robot to deviate from the stem; therefore, the state must be projected to maintain alignment. Each candidate position is projected onto the stem manifold by querying the $k$-d tree for the nearest traversable voxel centroid, with the successor state $s_{t+1}$ inheriting the pre-computed surface normal and branch direction at that point. Orientation consistency between states $s_t$ and $s_{t+1}$ is enforced via $\mathbf{n}_t \cdot \mathbf{n}_{t+1} > 0$ and $\mathbf{b}_t \cdot \mathbf{b}_{t+1} > 0$, where $\mathbf{n}_t$ and $\mathbf{b}_t$ are the normal and heading components of $s_t$, respectively. Vectors failing these conditions are inverted to maintain temporal consistency and prevent impossible motions caused by the sign ambiguity inherent in PCA-based eigen-decomposition.

\subsubsection{Collision Checking}

To ensure feasibility, candidate successor states are pruned if they result in collisions with geometry classified as obstacles, such as leaves or surrounding clutter. We maintain a $k$-d tree of the centroids of these voxels, denoted as the obstacle set $\mathcal{P}_{obs}$, for efficient spatial queries. A state $s_t$ is discarded if the Euclidean distance from the robot's position $\mathbf{p}_t$ to its nearest neighbor in $\mathcal{P}_{obs}$ is less than the robot's radius $r_{robot}$. Formally, a state is valid if:

$d(\mathbf{p}_t, \mathcal{P}_{obs}) = \min_{\mathbf{q} \in \mathcal{P}_{obs}} \|\mathbf{p}_t - \mathbf{q}\|_2 \geq r_{robot}$

\subsubsection{Branch Transition Constraints}
\begin{figure}[thpb]
    \centering
    \includegraphics[width=\linewidth]{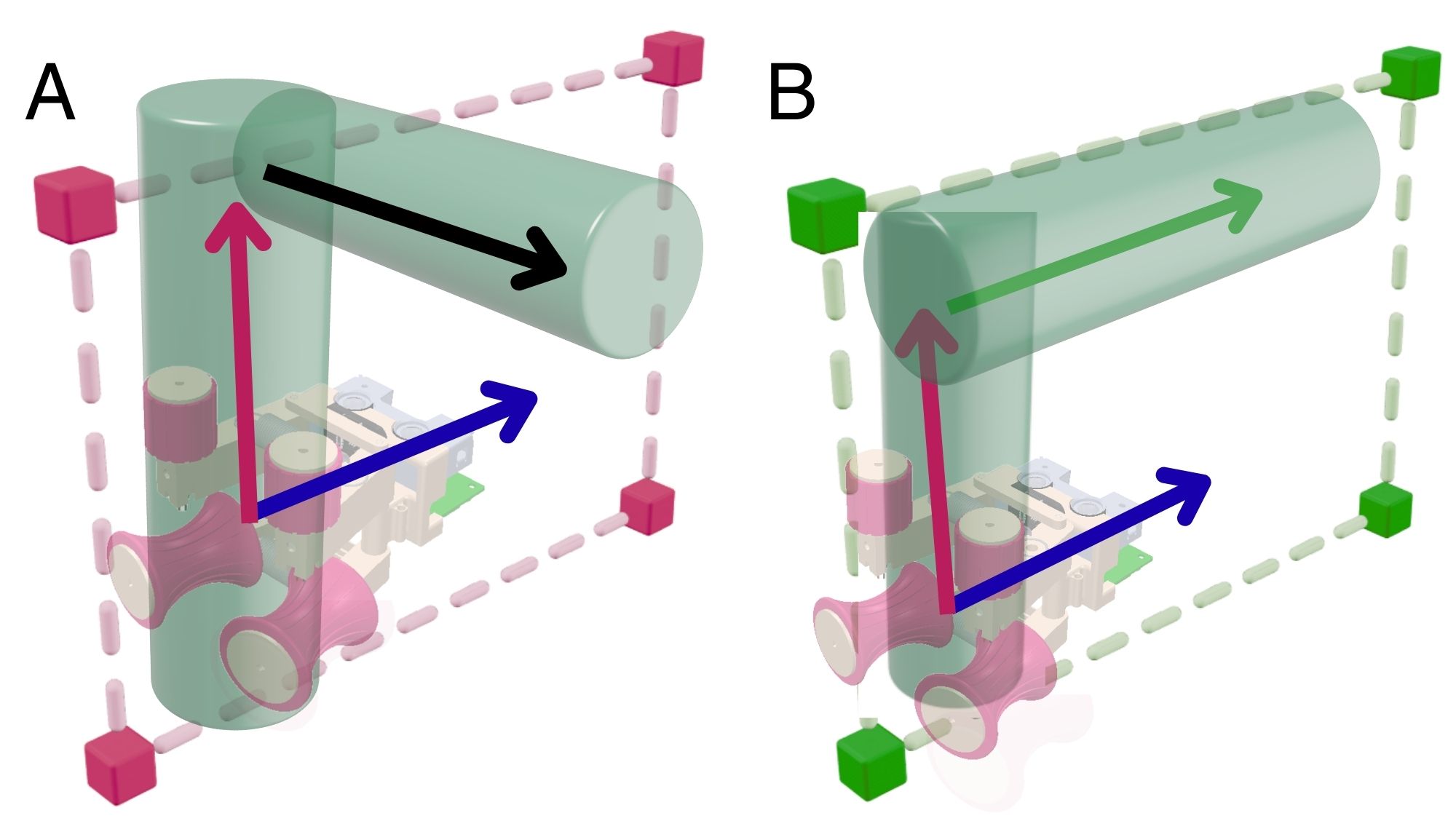}
    \caption{Geometric constraints for branch transition validation. (A) An invalid branch transition where the candidate successor is pruned from the A* search graph. The transition is rejected because the new branch vector $\mathbf{b}_{t+1}$ resides within the current plane defined by the branch and normal vectors ($\mathbf{b}_t, \mathbf{n}_t$), violating the docking geometry. (B) A valid branch switch where the candidate heading $\mathbf{b}_{t+1}$ satisfies the planarity and orientation constraints, indicating a feasible path through the junction that maintains continuous manifold contact.}
    \label{fig:branch_transition}
\end{figure}

Not all successor states are feasible due to docking constraints visualized in (Fig.~\ref{fig:branch_transition}); a potential branch switch is detected when $\mathbf{b}_t \cdot \mathbf{b}_{t+1} < \gamma$, where $\gamma$ is the branch detection threshold. Defining a transverse vector $\mathbf{m} = \mathbf{b}_t \times \mathbf{n}_t$, a transition is accepted only if it satisfies $|\mathbf{m} \cdot \mathbf{b}_{t+1}| < \epsilon$, where $\epsilon$ is the branch switching threshold. $\epsilon$ ensures that the successor heading $\mathbf{b}_{t+1}$ remains approximately within the plane defined by the current state, pruning transitions that require non-physical lateral maneuvers between stems. Invalid transitions are removed from the search graph (Fig.~\ref{fig:branch_transition}).

\begin{figure}[thpb]
    \centering
    \includegraphics[width=\linewidth]{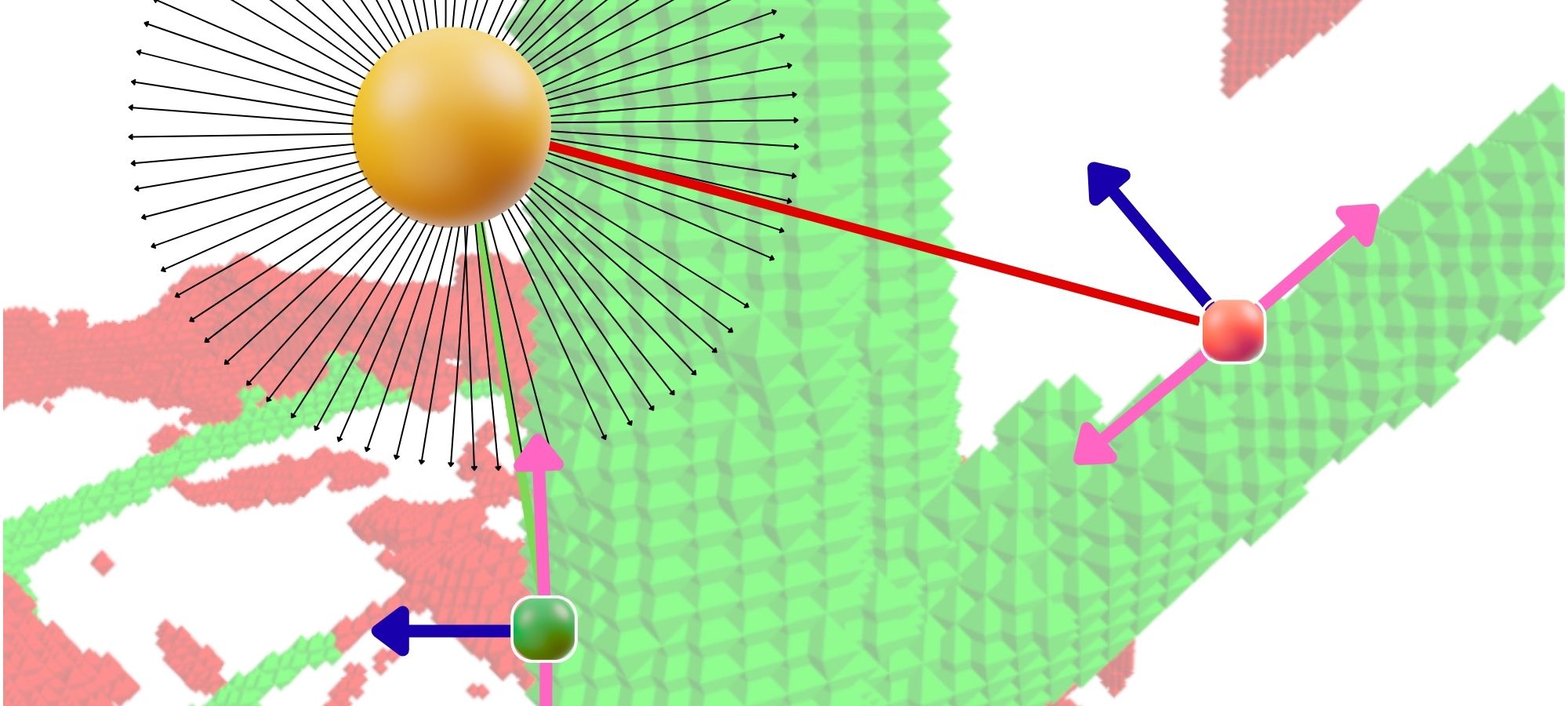}
    \caption{Goal state generation via radial ray-casting. Target point for observation is denoted in orange. Candidate viewing states are sampled using a Fibonacci sphere pattern; an example blue ray indicates a successful "hit" where the ray intersects the traversable manifold and the target resides within the camera frustum (which can point along either pink axis), resulting in a valid goal state. The red ray indicates a manifold intersection where the required robot orientation would place the target outside the camera frustum, causing the candidate state to be discarded.}
    \label{fig:goal_generation}
    \vspace{-15pt}
\end{figure}

\subsubsection{Cost Function and Heuristic}

The cost-to-come is defined as path length along the manifold:

\[
g(s_{t+1}) = g(s_t) + \|\mathbf{p}_{t+1} - \mathbf{p}_t\|_2.
\]

The heuristic $h(s)$ estimates the cost to the nearest goal state $s_g \in S_{\text{goals}}$ by combining spatial distance and heading alignment:$$h(s) = \min_{s_g \in S_{\text{goals}}} \left( \|\mathbf{p} - \mathbf{p}_g\|_2 + W_\theta \arccos(\mathbf{b} \cdot \mathbf{b}_g) \right)$$where $W_\theta$ weights the angular error. The set $S_{\text{goals}}$ contains either a single element for our state specification, or many possible states for the visibility goal specification.

\subsubsection{Goal Condition}

The planner supports two distinct goal modalities: state-based specification for waypoint navigation and visibility-constrained specification for targeted plant inspection. While the former directs the robot to a specific manifold coordinate, the latter defines the goal as any state $s$ from which a target point is both within the camera frustum and free from occlusion from leaves or stems.

The search terminates when all the following conditions are met:

\vspace{-15pt}
$$\|\mathbf{p} - \mathbf{p}_g\|_2 \leq d_{goal}, \quad \mathbf{n} \cdot \mathbf{n}_g > 0, \quad \text{and} \quad \mathbf{b} \cdot \mathbf{b}_g > 0,$$

where $d_{goal}$ represents the maximum allowable Euclidean distance between the current state position and the target goal. Additionally, the current surface normal $\mathbf{n}$ and branch heading $\mathbf{b}$ must align with their respective goal vectors, $\mathbf{n}_g$ and $\mathbf{b}_g$, which guarantees the robot is on the correct side of the branch and pointing in the intended direction.

Our visibility criterion  generates states from which a target point is visible, which is critical for tasks like pest inspection and exploration. Candidate viewing states are precomputed using radial ray-casting from the target with a Fibonacci sphere sampling pattern as shown in Fig. \ref{fig:goal_generation}. Traversable and obstacle clouds are represented as octahedral meshes for intersection testing.

For every ray that contacts a stem before an obstacle, a total of 4 states are added to a candidate set $S_{\mathrm{goals}}$, one for each possible combination of branch and normal vector directions at the contact point. Visibility of the target point is then checked using a fixed robot-to-camera transform and all valid states become A* goals. Planning is now a multi-goal search. Termination occurs when the planner reaches any candidate state within $d_{goal}$.

\subsubsection{Receding Horizon Re-planning}

In hardware deployment, stochasticity arises from wheel slippage, stem irregularities, and unmodeled environmental contact. To account for the discrepancy between planned trajectories and executed actions, we adopt a receding horizon framework. This approach mitigates model mismatch by re-computing the path at a high frequency. When the $A^*$ planner generates a sequence of action primitives and waypoints, only the first action in the sequence is dispatched to the robot controller, initiating a new $A^*$ search after the completion of each discrete action. By utilizing the updated odometry from PIN-SLAM in the new initial state $s_0$, the planner naturally compensates for any drift or obstacles encountered during the previous step. In cases of PIN-SLAM or mechanical failure, the trial is manually reset.

\section{RESULTS}

We evaluate our system on a series of experiments designed to validate mechanical robustness, mapping fidelity, and autonomous navigation. We characterize the robot's mechanical limits using 3D-printed geometries with varying stem diameters, curvatures, and bifurcations. We then assess the full-stack autonomy—integrating perception and manifold-constrained A* search—on both artificial and live plants. These trials employ both state and visibility specifications to demonstrate the versatility of our goal specification framework. Finally, we compare our 3D reconstructions against high-resolution photogrammetric ground truth to quantify the performance of our semantic mapping methods.

\subsubsection{Traversal Capability}

\begin{figure}[t]
    \vspace{10pt}
    \centering
    \includegraphics[width=\linewidth]{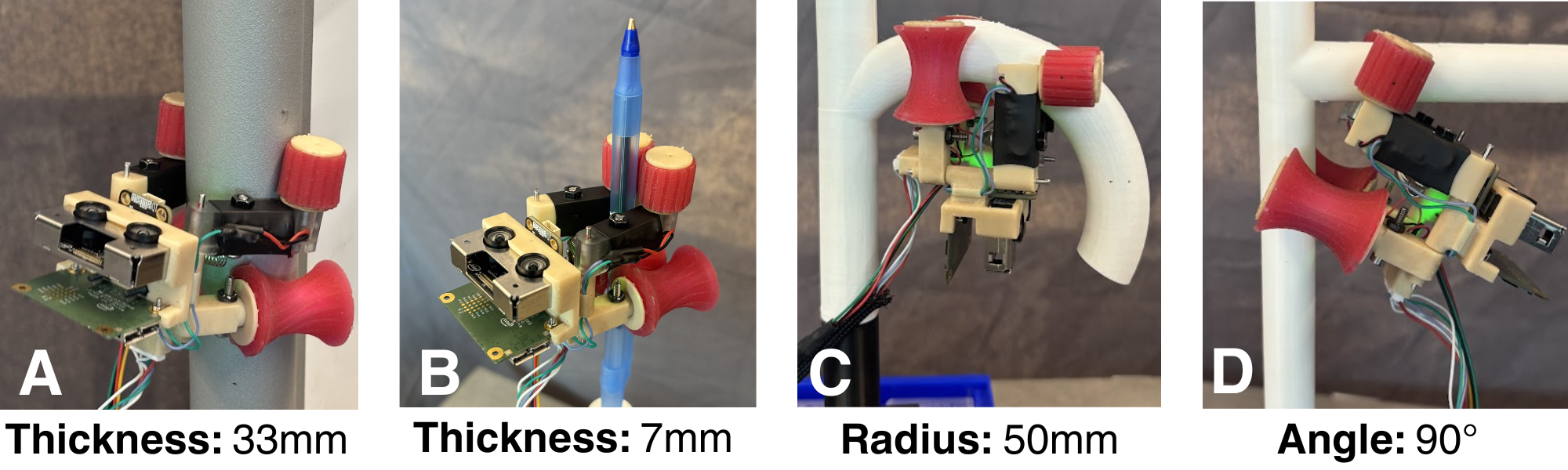}
    \vspace{-20pt}
    \caption{Traversal limits across stem geometries. (A)~33\,mm diameter whiteboard frame. (B)~7\,mm diameter Bic Round Stic pen. (C)~Curved branch with 50\,mm radius of curvature. (D)~90$^\circ$ branch junction.}
    \label{fig:traversal}
\end{figure}

To characterize the robot's mechanical limits, we evaluate its performance on a series of 3D-printed PLA benchmarks and common cylindrical objects. We isolate specific variables—stem diameter, stem bifurcation (branch) angle, and radius of curvature—which are non-uniform in plants. The robot successfully climbs stems ranging from a 7\,mm pen to a 33\,mm whiteboard frame (Fig.~\ref{fig:traversal}A--B), navigates branch junctions up to 90$^\circ$ on 16\,mm PLA stems (Fig.~\ref{fig:traversal}D), and traverses curving 20\,mm PLA branches with radii of curvature as tight as 50\,mm (Fig.~\ref{fig:traversal}C).

\begin{table}[t]
\caption{Parameter configurations across the experimental trials.}
\label{tab:parameters}
\centering
\addtolength{\tabcolsep}{-4pt}
\begin{tabular}{@{}lcccc@{}}
\toprule
 & \multicolumn{2}{c}{\textbf{Artificial}} & \multicolumn{2}{c}{\textbf{Live}} \\ \cmidrule(lr){2-3} \cmidrule(lr){4-5} 
\textbf{Parameter} & \textit{Drac.} & \textit{Olea} & \textit{Monst.} & \textit{Ficus} \\ \midrule
Goal Spec. & Point & Visib. & Visib. & Point \\
Goal Loc. $(x,y,z)$ & \scriptsize$(-.1,.3,1.5)$ & \scriptsize$(0,1,3)$ & \scriptsize$(0,0,2.5)$ & \scriptsize$(1,1,2.3)$ \\
ToF Dist. $d_{ToF}$ (mm) & 21 & 20 & 16 & 21 \\
Lin. Step $\delta_l$ (mm) & 0.1 & 0.1 & 0.1 & 0.2 \\ \midrule
\multicolumn{5}{c}{\textit{Global Parameters (All Trials)}} \\ \midrule
\multicolumn{5}{@{}c@{}}{
  \begin{tabular}{ll p{1em} ll}
  Ang. Step $\delta_a$ & 0.1$^\circ$ && Branch Switch Thresh. $\epsilon$ & 0.40 \\
  Rot. Weight $W_{\theta}$ & 1.0 && PCA $r_{normal}$ & 0.005 m \\
  Branch Detect Thresh. $\gamma$ & 0.995 && PCA $r_{branch}$ & 0.14 m \\
  Goal Dist. $d_{goal}$ & 0.01 m && & \\
  \end{tabular}
} \\ \bottomrule
\end{tabular}
\vspace{-20pt}
\end{table}

\subsubsection{Mapping Accuracy}

\begin{figure*}[t]
    \centering
    \includegraphics[width=\linewidth]{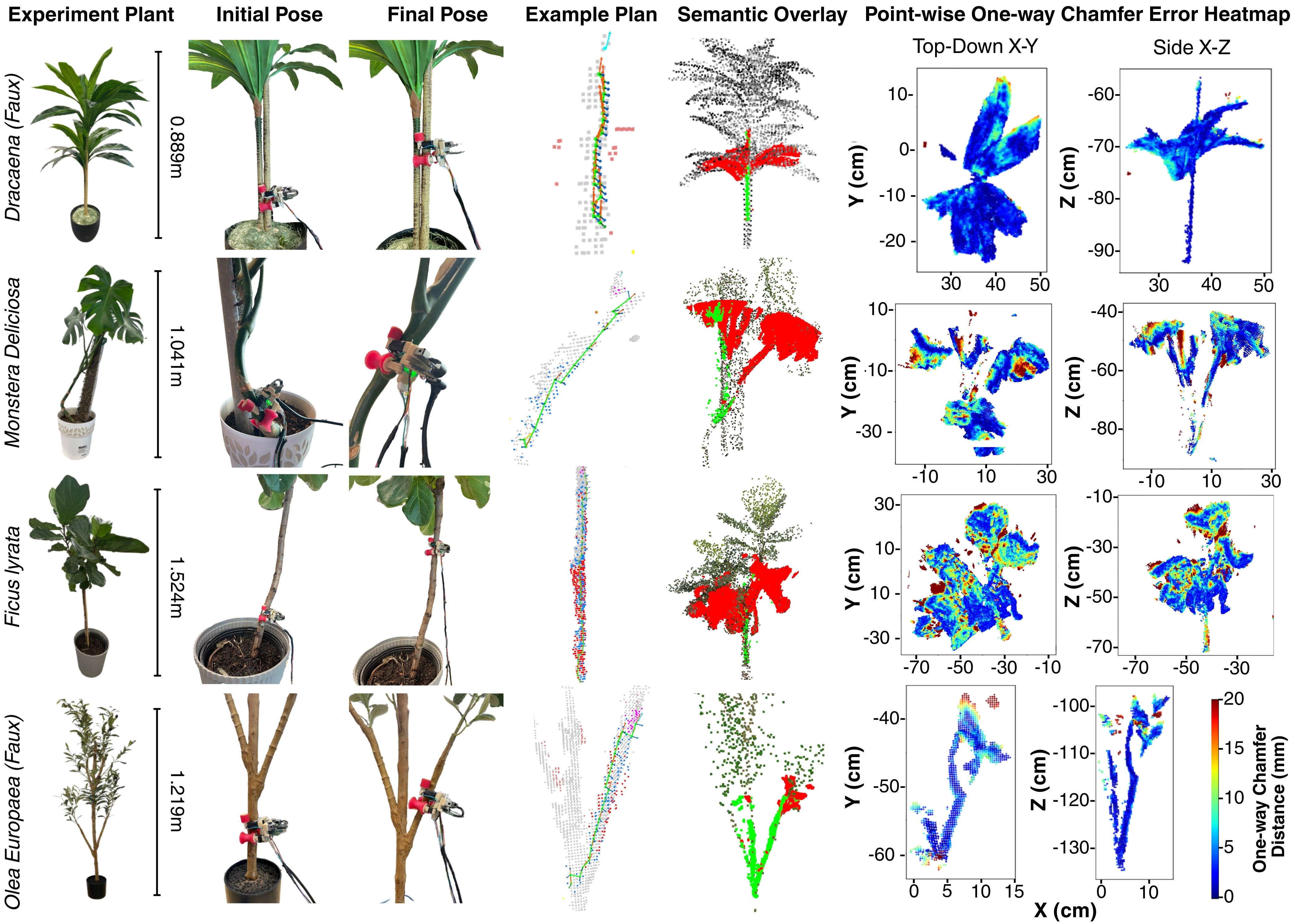}
    \vspace{-20pt}
    \caption{Experimental results for mapping and navigation across artificial and live specimens. Each row illustrates a trial, with columns representing (from left to right): the physical specimen with dimensions; initial and final robot poses; representative receding-horizon plans showing the path (green) and manifold geometry (normal $\mathbf{n}$ and branch $\mathbf{b}$ vectors in red/blue); qualitative semantic overlays; and point-wise one-way Chamfer error heatmaps.}
    \label{fig:mapping_experiments}
    \vspace{-15pt}
\end{figure*}

To ensure reliable navigation, the robot’s environment representation must remain globally consistent throughout its trajectory. We evaluate our mapping performance across four distinct specimens: two live (\textit{Monstera deliciosa}, \textit{Ficus lyrata}) and two artificial (\textit{Dracaena}, \textit{Olea europaea}), as illustrated in Fig.~\ref{fig:mapping_experiments}. In the Dracaena and Ficus trials, we used a state-based goal specification. For the Monstera and Olea europaea, we employed the visibility-constrained goal generation to inspect plant canopy regions that are initially occluded. Detailed parameter configurations for each experimental trial are summarized in Table~\ref{tab:parameters}. 

Fig.~\ref{fig:mapping_experiments} also illustrates representative trajectories generated by the receding horizon planner during each trial. Notably, the \textit{Olea europaea} plan captures a specific maneuver where the robot must reorient itself to satisfy docking constraints prior to transitioning between branches. Ground-truth (GT) geometry was established via an offline photogrammetry \cite{mikhail2001introduction} pipeline where 4K handheld video was processed using Agisoft Metashape (Standard Edition, Version 2.3) \cite{AgisoftMetashape2026} to produce dense point clouds. These were scaled with physical measurements and manually labeled and refined in CloudCompare (2.14.beta) \cite{Girardeau-Montaut2025}.

Figure~\ref{fig:mapping_experiments} illustrates the initial and final robot poses, qualitative semantic overlays, representative plans generated during execution, and point-wise error heatmaps. To assess the point-set error between the experimental reconstructions and GT scans, we utilize a one-way Chamfer distance. This metric compares the squared $L_2$ distance between each point in the reconstruction to its nearest neighbor in the GT, which is a standard approach for evaluating partial reconstructions. For the artificial specimens (Experiments 1 and 4), the system achieves a mean one-way Chamfer distance of $3.85\,\text{mm}$, demonstrating high reconstruction precision on rigid faux plants. In contrast, trials on live specimens (Experiments 2 and 3) exhibit a higher mean error of $13.36\,\text{mm}$. This variance is largely attributable to the non-rigidity of organic tissue and minor biological growth occurring between the baseline scan and experimental trials. Specifically, the ``Traversable" semantic class on real plants averaged an error of $37.56\,\text{mm}$, due in large part to the incorrect semantic classification of a primary branch on the Monstera (see Fig.~\ref{fig:mapping_experiments} semantic overlay). Segmentation of this specimen is particularly challenging due to low color contrast.

We observe an average Chamfer distance of 3.85\,mm on artificial plants and 13.36\,mm on live plants. We attribute this difference to violations of our static-environment assumption caused by slight branch deflections as well as plant growth and shape changes between ground truth scan collection and robot testing. The system maintained sufficient global map consistency to execute both state and visibility specifications. This suggests that integrating depth-based geometric SLAM with a manifold-aware state lattice offers a viable framework for navigating plant canopies.

\begin{figure}[thpb]
    \centering
    \includegraphics[width=\linewidth]{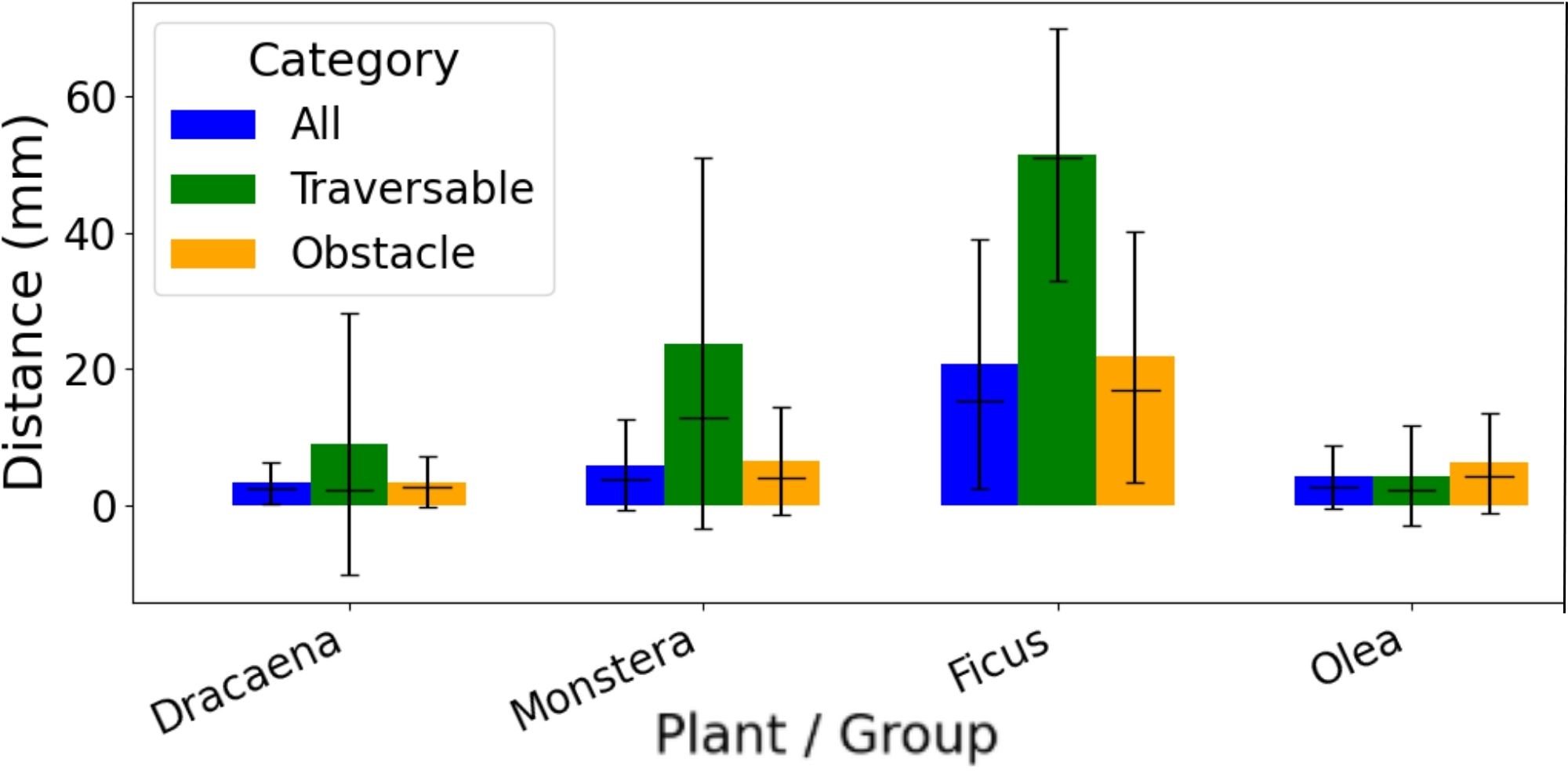}
    \label{fig:chamfer_metrics}
    \vspace{-20pt}
    \caption{Chamfer distance metrics (mean $\pm$ std, median) comparing experimental reconstructions to ground truth scans. The higher error observed in live trials highlights the combined impact of organic non-rigidity and semantic misclassification relative to artificial baselines.}
    \vspace{-20pt}
\end{figure}

\subsubsection{Failure Modes}

The results demonstrate the feasibility of autonomous navigation; however, future work is needed to address the following failure modes, which are exacerbated by the challenges of agricultural deployment.

\textit{Depth camera errors and noise.} Bright light exposure, especially direct sunlight, is a known challenge for RGB-D sensing systems. During testing, we turned off overhead lights for cleaner readings. Future filtering solutions will be need to address harsh direct sunlight.

\textit{PIN-SLAM tracking failure.} PIN-SLAM is prone to tracking loss when depth-sensing errors cause scan inconsistencies. Tracking loss can also occur if the robot moves too quickly relative to the PIN-SLAM update frequency. Once tracking is lost, camera-pose errors introduce map defects. The D401 depth module is much more accurate at close range; accordingly, the robot experienced fewer tracking-loss problems on the \textit{Dracaena} and \textit{Ficus lyrata} specimens, which have dense canopies and therefore produce STEMbot observations with fewer background pixels.

\textit{Hardware locomotion failures.} In some trials, bumpy protrusions on the \textit{Olea europaea} or \textit{Ficus lyrata}  required more torque to clear than our motors could provide. In other cases, a stem became wedged between the robot chassis and the silicone wheels, causing STEMbot to lose traction. Additionally, the ToF sensor can become occluded by a leaf, or may detect a knot in the wood, causing the robot to over or under-correct its pitch and lose contact with the branch.

\section{CONCLUSION}

We presented STEMbot, a miniature plant-climbing robotic system capable of navigation under plant canopies. By combining geometric PIN-SLAM odometry, probabilistic semantic octree mapping, and a manifold-aware A* planner, the system enables autonomous navigation while maintaining continuous contact with the stem. Our experiments demonstrate traversal across a range of stem geometries, accurate plant reconstruction, and the feasibility of closed-loop receding-horizon planning on real hardware.

Despite these results, several limitations remain. While our visibility goal specification enables the robot to reach a state where a specified target is observable, it does not yet reason about which regions of the plant are most informative to explore. Incorporating information-theoretic exploration strategies that explicitly optimize map or semantic uncertainty remains an important direction for future work. Additionally, our planning formulation assumes rigid plant geometry which we enforce by selecting stiff plants and using trellising in the case of the Monstera. Many real plants will bend and sway during traversal. Extending the perception and planning pipeline to account for compliant or dynamic plant structures is another key step toward robust deployment. Finally, improving robustness to failure modes and eliminating the need for a tether remain critical objectives for future work toward deployment in in-situ agricultural environments.

\bibliographystyle{IEEEtran}
\bibliography{IEEEabrv, bibliography}

\end{document}